\documentclass[letterpaper, 10 pt, conference]{IEEEconf}
\IEEEoverridecommandlockouts                              % This command is only needed if 
\usepackage{graphics} % for pdf, bitmapped graphics files
\usepackage{fontenc}
\usepackage{multirow}
\usepackage{graphicx}
\usepackage{subcaption}
\usepackage{tikz}
\usepackage[dvipsnames]{xcolor}
\usepackage{pgfplots}
\usepackage{amsfonts} 
\usepackage{amsmath} 
\usepackage{tikz-3dplot}
\usepackage{gensymb}

\title{\LARGE \bf
InsSo3D: Inertial Navigation System and 3D Sonar SLAM for turbid environment inspection
}

\author{Simon Archieri$^{1*}$, Ahmet Cinar$^{2}$, Shu Pan$^{1}$, Jonatan Scharff Willners$^{2}$, Michele Grimaldi$^{1}$, \\ Ignacio Carlucho$^{1}$ and Yvan Petillot$^{1}$ % <-this % stops a space
\thanks{$^{1}$School of Engineering and Physical Sciences, Heriot-Watt University, Edinburgh, UK}%
\thanks{$^{2}$Frontier Robotics, The National Robotarium, Edinburgh, UK}%
}

\begin{document}

\maketitle
\thispagestyle{empty}
\pagestyle{empty}

%%%%%%%%%%%%%%%%%%%%%%%%%%%%%%%%%%%%%%%%%%%%%%%%%%%%%%%%%%%%%%%%%%%%%%%%%%%%%%%%
\begin{abstract}
This paper presents InsSo3D, an accurate and efficient method for large-scale 3D Simultaneous Localisation and Mapping (SLAM) using a 3D Sonar and an Inertial Navigation System (INS). Unlike traditional sonar, which produces 2D images containing range and azimuth information but lacks elevation information, 3D Sonar produces a 3D point cloud, which therefore does not suffer from elevation ambiguity. We introduce a robust and modern SLAM framework adapted to the 3D Sonar data using INS as prior, detecting loop closure and performing pose graph optimisation. We evaluated InsSo3D performance inside a test tank with access to ground truth data and in an outdoor flooded quarry. Comparisons to reference trajectories and maps obtained from an underwater motion tracking system and visual Structure From Motion (SFM) demonstrate that InsSo3D efficiently corrects odometry drift. The average trajectory error is below 21cm during a 50-minute-long mission, producing a map of 10m by 20m with a 9cm average reconstruction error, enabling safe inspection of natural or artificial underwater structures even in murky water conditions.
\end{abstract}

% \begin{IEEEkeywords}
% SLAM, Marine Robotics, Mapping, Sonar
% \end{IEEEkeywords}

%%%%%%%%%%%%%%%%%%%%%%%%%%%%%%%%%%%%%%%%%%%%%%%%%%%%%%%%%%%%%%%%%%%%%%%%%%%%%%%%
\section{INTRODUCTION}
\begin{figure}
    \centering
    \includegraphics[width=.9\linewidth]{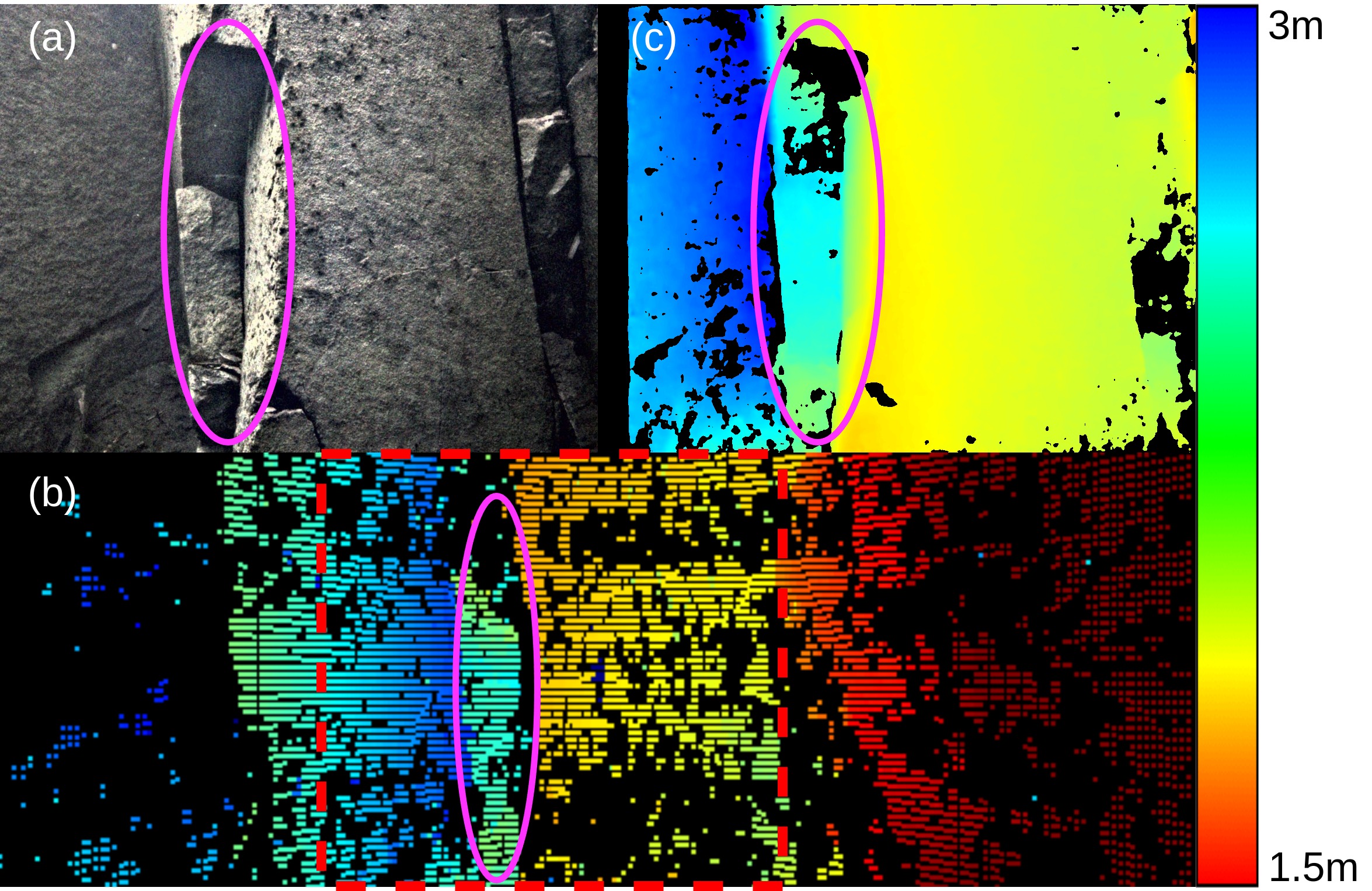}
    \caption{Comparison between (a) Camera image, (b) 3D Sonar depth image, and (c) Camera depth image generated with RoMa~\cite{edstedt2024roma} thanks to excellent visibility conditions. The red rectangle indicates the overlapping area between 3D Sonar and camera FOV, and the pink ellipse highlights a similar feature. The range colour scale is indicated on the right. Also, the camera FOV is 60\textdegree~by 45\textdegree, and the 3D Sonar FOV is  90\textdegree~by 40\textdegree.}
    \label{fig:sonar3d_example}
\end{figure}

Underwater 3D perception is an essential component of safe navigation in an unknown environment for Autonomous Underwater Vehicles (AUVs), and critical for many tasks from underwater mapping to offshore asset inspection. Most existing approaches use optical solutions with a single or multiple cameras \cite{ORBSLAM3_TRO} or based on lidar \cite{filisetti2018developments}. Although these solutions are effective, they are very dependent on water conditions and work only at short ranges, generally less than a few meters. Acoustic solutions, such as Sonar, on the other hand, are less dependent on water conditions and can operate at much larger ranges. However, their signal-to-noise ratio and spatial resolution are inferior to those of an optical-based system. Acoustic imaging is also subject to a wide variety of complex noise and multipaths, which makes automatic processing very challenging. Traditional sonar covers a fan-shaped Field of View (FOV) and produces 2D images representing a projection of the volume of water observed, where the elevation angle is lost. Resolving the elevation ambiguity is very challenging and is the main reason why most perception solutions using sonar have been limited to 2D \cite{bslam}. However, innovations from sonar manufacturers have enabled the creation of a new generation of sonar which does not suffer from this elevation ambiguity; we refer to those sonars as 3D Sonar. Those new sonars generate a depth images akin to a lidar where each pixels are characterised by its range, azimuth and elevation angles. 3D Sonar are also affected by the different types of noise mentioned above and generally produces a sparse measurement, making automatic registration of 3D Sonar frames challenging. Fig. \ref{fig:sonar3d_example} shows a comparison of 3D Sonar and Camera data.

\begin{figure}
    \centering
    \includegraphics[width=.9\linewidth]{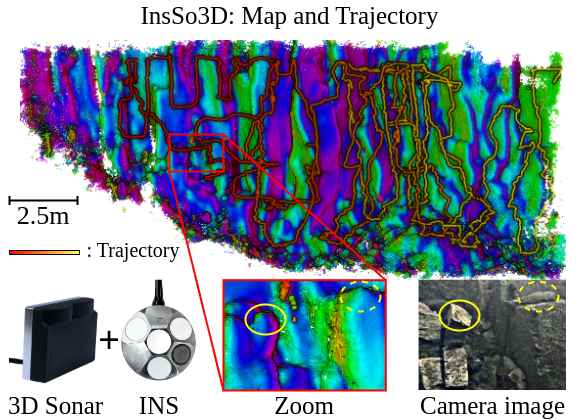}
    \caption{InsSo3D uses the 3D Sonar and the INS to estimate the robot's trajectory and generate a 3D map of its environment. The trajectory colour indicates the acquisition timestamp of the 3D Sonar frame.}
    \label{fig:example}
\end{figure}

In this article, we address the challenge of Simultaneous Localisation and Mapping (SLAM) using a 3D Sonar as shown in Fig. \ref{fig:example}. To achieve this, we rely solely on a 3D Sonar and an INS composed of: a Doppler Velocity Log (DVL), an Attitude and Heading Reference System (AHRS) and a pressure sensor. 

We evaluated our method in real-world experiments by comparing its results to photogrammetry data obtained from a stereo camera and using a motion tracking system. The average trajectory error relative to the reference remains consistently below $21cm$. Furthermore, our method achieves large-scale SLAM, reconstructing a $20m \times 10m$ environment with an average map error of $9cm$ compared to the reference map during our longest experiment. Those results demonstrate the effectiveness of the method for accurate SLAM, effectively correcting odometry drift and generating a large and consistent map. The key contributions enabling this performance are:
\begin{itemize}
    \item Use of a 3D Sonar that enables robust, large-scale, and accurate SLAM of an unstructured environment.
    \item Integration of CFEAR~\cite{9969174} for robust, noise-resistant 3D frame and sub-map matching, loop closures detection and global alignment.
    \item Comprehensive validation in different real-world underwater environments with quantitative performance analysis.
\end{itemize}

\section{Related Work}
Underwater 3D mapping and SLAM systems are fundamentally shaped by the sensing modality employed, as underwater environments impose unique challenges not present in terrestrial or aerial domains. These include limited visibility due to turbidity, light absorption with depth, dynamic water currents, and complex acoustic conditions such as multipath and scattering \cite{grimaldi2023investigation, foote2018using}. Consequently, the choice of sensor affects not only the fidelity and scale of generated maps but also the feasibility of real-time localisation, long-term autonomy, and operational robustness. The two dominant classes of sensors in this domain are optical sensors, which leverage light to capture high-resolution visual or depth information, and acoustic sensors, which rely on sound propagation to probe the environment.

Optical sensors are widely used for underwater SLAM due to their ability to capture high-resolution textures and geometric features \cite{johnson2010generation}. Monocular cameras provide lightweight, low-power solutions suitable for small autonomous underwater vehicles (AUVs), with SLAM techniques adapted from ORB-SLAM \cite{ORBSLAM3_TRO, 8559435} recovering scene structure through feature tracking and temporal geometry estimation. However, monocular methods suffer from scale ambiguity and rely on external cues, such as pressure sensors, to resolve metric scale. Their performance also degrades in visually challenging environments, such as turbid water or featureless seafloors \cite{grimaldi2023investigation}. Stereo camera systems attempt to overcome some of these limitations by triangulating depth from two calibrated cameras, enabling dense point-cloud mapping and detailed 3D reconstruction. While successful in underwater inspection, stereo systems remain sensitive to turbidity, lighting variation, and backscatter, require precise calibration, and are limited in effective range. These constraints restrict their applicability for large-scale, deep, or highly dynamic underwater environments. 

\begin{figure*}[]
    \centering
    \includegraphics[width=.9\textwidth]{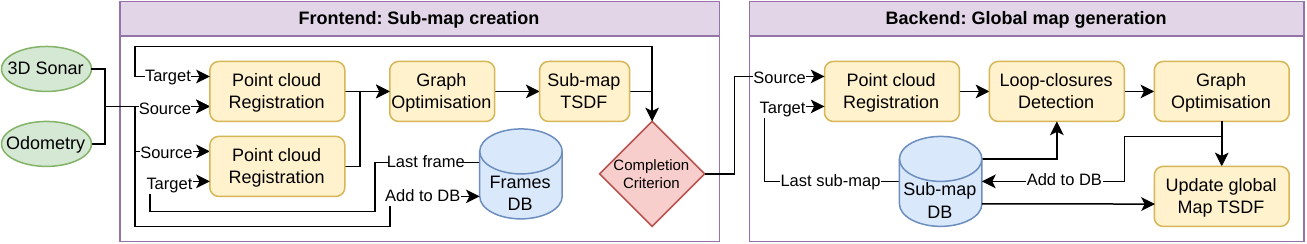}
    \caption{Algorithm overview: new 3D Sonar frames are registered to the current sub-map using frame-to-frame and frame-to-sub-map registration. Then, a graph optimisation within the sub-map is performed to fuse the two registrations and estimate the correct frame pose. Finally, the new frame is integrated into the sub-map TSDF. When the submap completion criterion is met, the submap is closed and sent to the backend. The backend will register it to the previous one and try to detect loop closures with older sub-maps contained in the sub-map database (DB) before including the new sub-map in the factor graph. Finally, a TSDF representation of the global map is generated using each sub-map TSDF representation and the sub-map poses optimised by the factor graph.}
    \label{fig:overview}
\end{figure*}

In contrast, acoustic sensors like sonars remain the backbone of underwater perception, operating effectively where optical systems fail, such as in turbid, low-light, or deep-sea environments. By exploiting long-range sound propagation, sonars enable robust mapping and navigation in visually degraded settings. Depending on type and configuration, acoustic sensors can support 2D SLAM and 3D reconstruction, typically at the cost of lower spatial resolution, slower updates, and higher processing demands. Traditional sonar systems cover a fan-shaped FOV but only provide 2D images with range and bearing information. This lack of elevation information is a central challenge in underwater perception with imaging sonar. Addressing this ambiguity is therefore critical for building an accurate 3D representation. Four main categories of SLAM using 2D sonar stand out. 
Early work has largely sidestepped elevation ambiguity by projecting the 2D sonar image onto a 2D horizontal plane~\cite{xu2024diso, zhuang2024graph, bslam, pan2025russo} or projecting the 2D detected features along the 3D elevation arc to help improve the 2D localisation~\cite{bai2025sio, archieri20253dssdf}. However, scan matching in 2D projections is fundamentally limited, often leading to significant errors in environments with any elevation variation, a condition almost always met in underwater applications.
Second, classical methods address the elevation ambiguity problem primarily through geometric modelling and multi-view integration ~\cite{aykin20153,teixeira2019dense, huang2015towards, guerneve2018three, wang2020acoustic}. These approaches can provide accurate reconstructions when vehicle motion is well-constrained and navigation data is reliable. However, classical methods are highly sensitive to degenerate motion cases such as pure translation or pure rotation, where elevation ambiguity cannot be resolved~\cite{huang2015towards}. This requirement for well-constrained motion makes piloting and trajectory planning more challenging in practice.
A promising solution to the elevation ambiguity is the use of multiple orthogonally mounted imaging sonars, enabling direct triangulation of reflective targets in 3D, analogous to stereo vision \cite{sadjoli2023pcd, mcconnell2024large, liu2024target}, overcoming the fundamental limitation of single-sonar systems. However, the practical deployment of orthogonal sonar configurations faces several challenges. The overlapping field of view between sensors is often limited, resulting in sparse coverage and incomplete reconstructions. Additionally, the increased size, complexity, and cost associated with multiple sonars may be prohibitive.
Finally, learning-based approaches are also used for solving the elevation ambiguity~\cite{debortoli2019elevatenet,jaber2023sonar2depth}, leveraging neural networks to extract 3D information directly from acoustic images. However, such methods depend on large labelled datasets, which are costly to collect underwater, limiting scalability and generalisation. 
A major leap in sonar 3D reconstruction has come from neural implicit and neural rendering methods~\cite{gao2025lambertian,qadri2022neuralimplicitsurfacereconstruction} inspired by Neural Radiance Fields (NeRF) \cite{mildenhall2021nerf}. However, most neural implicit approaches are computationally intensive, typically requiring minutes to hours for reconstruction, and are suited for offline processing. Moreover, they remain sensitive to pose accuracy; uncorrected navigation drift can degrade reconstruction quality.
Although traditional sonar SLAM has demonstrated strong performance, extending it to full 6DOF estimation remains a significant challenge as it usually requires computationally intensive methods and highly constrained motion to resolve scene geometry, largely due to elevation ambiguity. In contrast, 3D sonar SLAM inherently supports 6DOF pose estimation, as it is not affected by elevation ambiguity, enabling more robust and flexible performance in complex underwater environments. Up to now, very few 3D sonars have been available, and they have all been expensive, bulky, and power-hungry (such as the echoscope), making them unsuitable for 3D SLAM onboard small platforms. Their use in SLAM applications has been limited—primarily due to prohibitive costs. Consequently, the few existing works on 3D sonar SLAM have focused mainly on frame-to-frame matching~\cite{ferreira2025real,hansen2005mosaicing}. These approaches do not incorporate loop closure detection or sub-maps, which are essential for global consistency and long-term mapping. However, recent innovations have significantly reduced the cost of 3D sonar systems, paving the way for more accessible and scalable SLAM implementations, including the development presented in this work.

To overcome these limitations, we propose a SLAM system based on the \textit{Waterlinked Sonar3D-15}, a small and low-power volumetric acoustic sensor capturing azimuth, elevation, and range in a single ping. Unlike conventional approaches, it produces depth images in real time without post-processing, as can be seen in Fig. \ref{fig:sonar3d_example}. This direct sensing enables robust, low-latency 3D mapping and navigation in dark, turbid, or turbulent conditions, offering a compact, high-reliability solution for operationally constrained underwater environments.

\section{METHOD}

In this section, we present the key components of the methodology. As shown in Fig. \ref{fig:overview}, the algorithm is split into two parts. First, the Sonar 3D frames are accumulated into a sub-map by using a frame-to-sub-map and frame-to-frame registration. When the sub-map completion criterion is met, the new sub-map is closed and registered to those captured previously to create a consistent global map. We will first present the point cloud registration algorithm that uses the odometry as prior. This algorithm is used for frame-to-frame, frame-to-sub-map and sub-map to sub-map registration. Second, we will introduce the frontend algorithm responsible for sub-map creation and finally, how the backend registers and merges new sub-maps with older ones in order to build a consistent global map using the Truncated signed distance function (TSDF) approach. 

\subsection{Notations}
Before we start with the description of our method, we outline the notation used in this section to ensure clarity for readers. The odometry frame is depicted as $\mathcal{F}_o$ in NED coordinates, while $\mathcal{F}_f$ represents the position of the robot in the frontend frame after correction by the frontend but before backend correction. The world frame representing the robot position after correction by the frontend and the backend is depicted as $\mathcal{F}_w$.
Thus a 3D Sonar frame $f$ is characterised by its pose in the odometry frame $\mathcal{F}_o$ and in the frontend frame $\mathcal{F}_f$, denoted as $^o\mathbf{F}(f)$ and $^f\mathbf{F}(f)$ respectively. Similarly, a sub-map $s$ is characterised by its pose in the odometry frame $\mathcal{F}_o$ and in the corrected world frame $\mathcal{F}_w$, denoted as $^o\mathbf{P}(s)$ and $^w\mathbf{P}_n(s)$ respectively. Note that each pose $^w\mathbf{P}_n(s)$ of a specific sub-map in $\mathcal{F}_w$ is kept after each graph optimisation and represents the pose of the sub-map after optimisation number $n$. This way, we keep a history of the different sub-maps pose to adjust the global map TSDF if needed after graph optimisation. In practice, we optimise the graph after acquiring a new sub-map, and therefore $n$ corresponds to the total number of sub-maps. 

\subsection{Point cloud registration} \label{section:pc_reg}

\begin{figure}[]
    \centering
    \begin{subfigure}{0.25\columnwidth}
        \includegraphics[width=\textwidth]{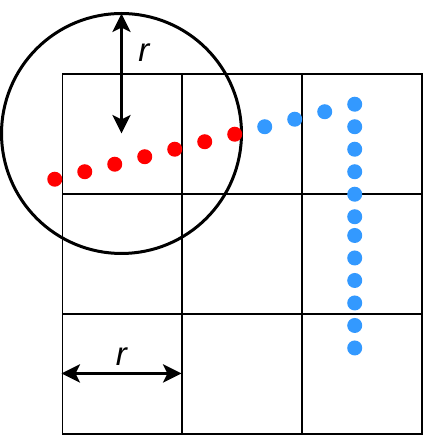}
        \caption{}
        \label{fig:cfear:1}
    \end{subfigure}
    \begin{subfigure}{0.25\columnwidth}
        \includegraphics[width=\textwidth]{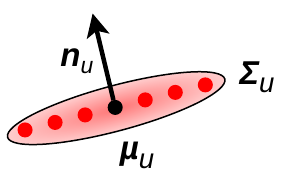}
        \caption{}
        \label{fig:cfear:2}
    \end{subfigure}
    \begin{subfigure}{0.25\columnwidth}
        \includegraphics[width=\textwidth]{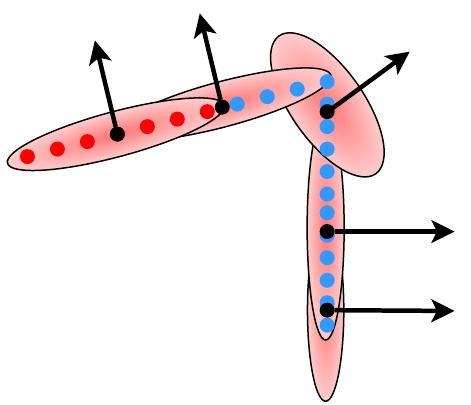}
        \caption{}
        \label{fig:cfear:3}
    \end{subfigure}
    \caption{Process of computing the \textit{CFEAR} oriented surface point cloud. (a) The point cloud is inserted into a voxel grid of voxel size $r$. (b) The mean and covariance are computed for each voxel using points within a distance $r$ of the voxel centroid. (c) Final representation.}
    \label{fig:cfear}
\end{figure}

A 3D variant of \textit{CFEAR} \cite{9969174} is used to register two point clouds together. \textit{CFEAR} generates a sparse representation for both the source and target point clouds. This representation aims to model the scene geometries as a set of oriented surface points that are stable in at least one direction. First, as shown in Fig. \ref{fig:cfear}(a), the point cloud is inserted into a voxel grid of resolution $r$. For each voxel $u$, all points within a radius $r$ from the voxel centroid are used to compute the sample mean $\mu_u$ and covariance $\Sigma_u$ of the voxel as shown in Fig. \ref{fig:cfear}(b) and \ref{fig:cfear}(c). By applying the eigen decomposition of $\Sigma_u$, the surface normal $n_u$ can be estimated using the eigenvector that corresponds to the smallest eigenvalue. The oriented surface points are therefore defined by $(\mu_u, n_u)$ for each voxel $u$ that has more than 6 points within a sphere of radius $r$.

Unlike \textit{CFEAR}, we use \textit{fast\textunderscore gcip} \cite{EasyChair:2703}, an implementation of Generalized ICP \cite{inproceedings} to optimise the registration transformation between the source and target oriented surface point clouds. Generalized ICP uses a plane-to-plane distance metrics, which make it suitable for oriented surface point-cloud registration. The maximal correspondence distance of GICP is set to $r$, the intuition is that for aligned scans, nearby surface points are expected to lie within this distance. 

\subsection{Frontend: Sub-map Creation}
\subsubsection{Registration}

\begin{figure}[t]
    \centering
    \includegraphics[width=0.7\linewidth]{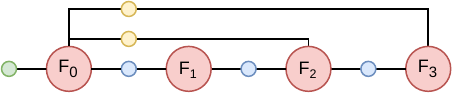}
    \caption{Factor graph representation of the \textbf{frontend}, \textbf{red} nodes are the frame's pose, \textbf{blue} nodes are sequential frame-to-frame factors, and the \textbf{yellow} nodes are frame-to-sub-map factors. Finally, the \textbf{green} node is a prior factor.}
    \label{fig:frontend:graph}
\end{figure}

A new incoming 3D Sonar frame $i$ is projected into $\mathcal{F}_f$ using the relative motion from the odometry between frames $i$ and $i-1$ using the following equation:
\begin{equation}
    ^f\mathbf{F}(i)^* = \begin{cases}
    ^f\mathbf{F}(i-1) \ {^o\mathbf{F}(i-1)^{-1}}\  {^o\mathbf{F}(i)}, & \text{if $i>0$}.\\
    ^o\mathbf{F}(i), & \text{otherwise}.
    \end{cases}
\end{equation}
In order to increase robustness (see Fig. \ref{fig:overview}), InsSo3D performs two types of registration within the frontend. First, a frame-to-frame registration is performed between frames $i$ and $i-1$. Second, the new frame $i$ is registered either to the TSDF point cloud representation of the sub-map or to the first frame $i^*$ of the sub-map, depending on which one has the highest number of points. 
After performing both registrations, the new frame and its registrations are added to the frontend factor graph as shown in Figure \ref{fig:frontend:graph}. Frame-to-frame registrations are included using a sequential between-factor between frame $i$ and $i-1$. Frame-to-sub-map registrations are included using a non-sequential between-factor connecting the new frame $i$ to the first sub-map frame $i^*$. The pose $^f\mathcal{F}(i)$ is therefore estimated during graph optimisation.

\subsubsection{Sub-map TSDF}
After graph optimisation the new 3D Sonar frames are integrated into the sub-map TSDF using the space carving variation introduced in \cite{vizzo2022sensors}. Compared to normal TSDF, where only voxels inside the truncation region are updated when integrating a point, TSDF space-carving updates all voxels along the integration ray. With TSDF space-carving, each voxel has a maximum absolute value equal to the truncation threshold. This variation allows for keeping track of regions that have been observed and considered free. We generate a TSDF model for each sub-map independently before any backend correction so that they can be rigidly transformed if needed after backend graph optimisation. 
Moreover, as the frontend registration needs a real-time point cloud representation of the TSDF, the use of a meshing algorithm like \textit{Marching Cube} \cite{lorensen1998marching} is not convenient. Instead, a coarse point cloud representation is generated by converting to a point each voxel that has an absolute signed distance below half of the voxel size~\cite{archieri20253dssdf}.

\subsubsection{Sub-map completion criterion}
Sub-maps are considered to be complete when the percentage of overlap between the last and first 3D Sonar frame is below a threshold. The intuition is to keep sub-maps small to limit the pose drift inside the sub-map, as no loop closures are being detected within the sub-map. In addition, the sub-map TSDF point cloud representation must contain more points than the first sub-map frame to be considered complete. This avoids closing sub-maps that are too small or in the event of a wrong frame-to-sub-map registration due to insufficient points in the registration target. In this event, the frontend can recover from the bad registration, as we also use frame-to-frame registration.

The pose of a sub-map $s$ in $\mathcal{F}_o$ is defined by the pose of the first 3D Sonar frame $i^*$ contained in the sub-map as $^o\mathbf{P}(s)=^o\mathbf{F}(i^*)$.

\subsection{Backend: Global Map Generation}
\subsubsection{Registration} \label{section:graph}

\begin{figure}[]
    \centering
    \includegraphics[width=0.7\linewidth]{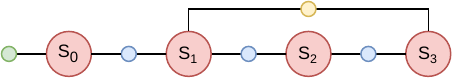}
    \caption{Factor graph representation of the \textbf{backend}, \textbf{red} nodes are the sub-map pose, \textbf{blue} nodes are sequential sub-maps to sub-map factors, and the \textbf{yellow} node is a loop closure factor. Finally, the \textbf{green} node is a prior factor.}
    \label{fig:backend:graph}
\end{figure}

A pose-graph framework is used within the backend to optimise the sub-map pose \cite{factor_graphs_for_robot_perception}. When a new sub-map $k$ is generated it is first projected into $\mathcal{F}_w$ using the relative motion from the odometry between sub-map $k$ and $k-1$ using the following equation:
\begin{equation}
    ^w\mathbf{P}(k)^* = \begin{cases}
    ^w\mathbf{P}_{k-1}(k-1) \ {^o\mathbf{P}(k-1)^{-1}}\  {^o\mathbf{P}(k)}, & \text{if $k>0$}.\\
    ^o\mathbf{P}(k), & \text{otherwise}.
    \end{cases}
\end{equation}
A sequential sub-map to sub-map registration is then performed between sub-map $k$ and $k-1$. Therefore the sub-map $k$ position in $\mathcal{F}_w$ is refined as $^w\mathbf{P}(k)^{**}$ using the following equation:
\begin{equation}
    ^w\mathbf{P}(k)^{**} = {^{k-1}\mathbf{T_{ICP}}_k}  {^w\mathbf{P}(k)^*}.
    \label{eq:graph:temp_lc}
\end{equation}
Where ${^{k-1}\mathbf{T_{ICP}}_k}$ is the estimated transformation obtained using the registration algorithm. This refined pose is then used to detect loop-closures as explained in the section \ref{section:lc}.
Those calculated transformations are then inserted into a factor graph as shown in Fig. \ref{fig:backend:graph}. The pose $^w\mathbf{P}_n(k)$ is therefore obtained after the graph optimisation.

\subsubsection{Loop closures} \label{section:lc}

When the robot revisits a previously explored part of its environment, a loop-closure constraint can be added to the factor graph. In order to detect those loop closures, when a new sub-map $k$ is being registered, a set of sub-maps with potential loop closures $\mathcal{S}_{lc}(k)$ is identified. To do so, the new sub-map $k$ point cloud representation is inserted into a voxel grid of resolution $r$. Then, every other sub-map point cloud representations are inserted into different voxel grids of the same resolution, and their overlap ratio with the new sub-map $k$ voxel grid is calculated. The corresponding sub-map is added to $\mathcal{S}_{lc}(k)$ if the overlap ratio exceeds $50\%$. Afterwards, the sub-map $k$ is registered with each sub-map $s \in \mathcal{S}_{lc}(k)$. If the final registration error between sub-map $k$ and sub-map $s$ is below a threshold, a loop closure will be considered during graph optimisation. This loop closure will use the transformation calculated during the registration.

\subsubsection{Global Map TSDF} \label{section:gtsdf}

In order to adapt the global map accordingly to the backend graph optimisations, sub-maps are integrated into TSDF independently and then merged together. Each sub-map that has moved after graph optimisation is deleted from the global map and then re-added with a corrected pose. To achieve this, InsSo3D saves $\phi_n(k)$, the index at which a sub-map $k$ is integrated into the global map, as graph optimisation $n$ is made. In practice for real-time implementation, a sub-map $k$ will be reprocessed only if the rotation and the translation between $^w\mathbf{P}_n(k)$ and $^w\mathbf{P}_{\phi_{n-1}(k)}(k)$ are above a threshold. 

Let $D_n(\mathbf{x})$ and $W_n(\mathbf{x})$ be the SDF and weight function of the global map at graph optimisation $n$ in $\mathcal{F}_w$. Furthermore, $d_k(\mathbf{x})$ and $w_k(\mathbf{x})$ are the equivalents for the sub-map $k$ in $\mathcal{F}_o$, and $\mathcal{S}_{m}(n)$ is a set containing the sub-map whose pose moved during the graph optimisation step $n$. The updated global map SDF and weight function considering the new sub-map $n$ and the moved sub-map in $\mathcal{S}_{m}(n)$ are expressed as follows:
\begin{equation}
    D_n(\mathbf{x}) = \frac{W_{n-1}(\mathbf{x})D_{n-1}(\mathbf{x}) + \psi_{n,n}(\mathbf{x}) + U(\mathbf{x})}
    {W_n(\mathbf{x})},
\end{equation}
\begin{equation}
    W_n(\mathbf{x}) = W_{n-1}(\mathbf{x}) + w_n(\theta_{n,n}(\mathbf{x})) + UW(\mathbf{x}).
\end{equation}
$D_n(\mathbf{x})$ is the sum of three terms: the existing value $W_{n-1}(\mathbf{x})D_{n-1}(\mathbf{x})$, the new sub-map $n$, $\psi_{n,n}$, and the changes caused by the sub-maps whose pose have moved $U(\mathbf{x})$. Similarly, $W_n(\mathbf{x})$ is the sum of the three equivalent terms for the new weight calculation. $U(\mathbf{x)}$ and $UW(\mathbf{x})$ are calculated for each sub-map $k\in\mathcal{S}_m$ by removing the sub-map using the pose ${^wP_{\phi_{n-1}}(k)}$ and adding it back using the pose ${^wP_n(k)}$, as in the following equations:
\begin{equation}
    U(\mathbf{x}) = \sum\limits_{k \in \mathcal{S}_{m}(n)} \psi_{n,k}(\mathbf{x}) - \psi_{\phi_{n-1}(k),k}(\mathbf{x}),
\end{equation}
\begin{equation}
    UW(\mathbf{x}) = \sum_{k \in \mathcal{S}_m(n)} w_k(\theta_{n,k}(\mathbf{x})) - w_k(\theta_{\phi_{n-1}(k),k}(\mathbf{x})).
\end{equation}
Where $\psi_{q,k}(\mathbf{x})$ is the weighted signed distance in $\mathcal{F}_w$ of the sub-map $k$ using the pose at graph optimisation $q$. Finally, the function $\theta_{q,k}(\mathbf{x})$ transforms a point $\mathbf{x}$ of a sub-map $k$ from $\mathcal{F}_w$ to $\mathcal{F}_o$ using the pose at graph optimization $q$. This transformation is required because the TSDF representations of each sub-map are independently integrated in $\mathcal{F}_o$.
\begin{equation}
    \psi_{q,k}(\mathbf{x}) = w_k(\theta_{q,k}(\mathbf{x})) \times d_k(\theta_{q,k}(\mathbf{x})),
\end{equation}
\begin{equation}
    \theta_{q,k}(\mathbf{x})= {^o\mathbf{P}(k)} \times {^w\mathbf{P}_{q}(k)^{-1}}.
\end{equation}
Finally, as the global map rendering is not needed in real-time, the \textit{Marching Cube} algorithm \cite{lorensen1998marching} is used to generate a mesh representation of the global map TSDF.

\section{EVALUATION} \label{section:eval}

\begin{figure}[b]
    \centering
    \begin{subfigure}{0.45\columnwidth}
        \includegraphics[width=\textwidth]{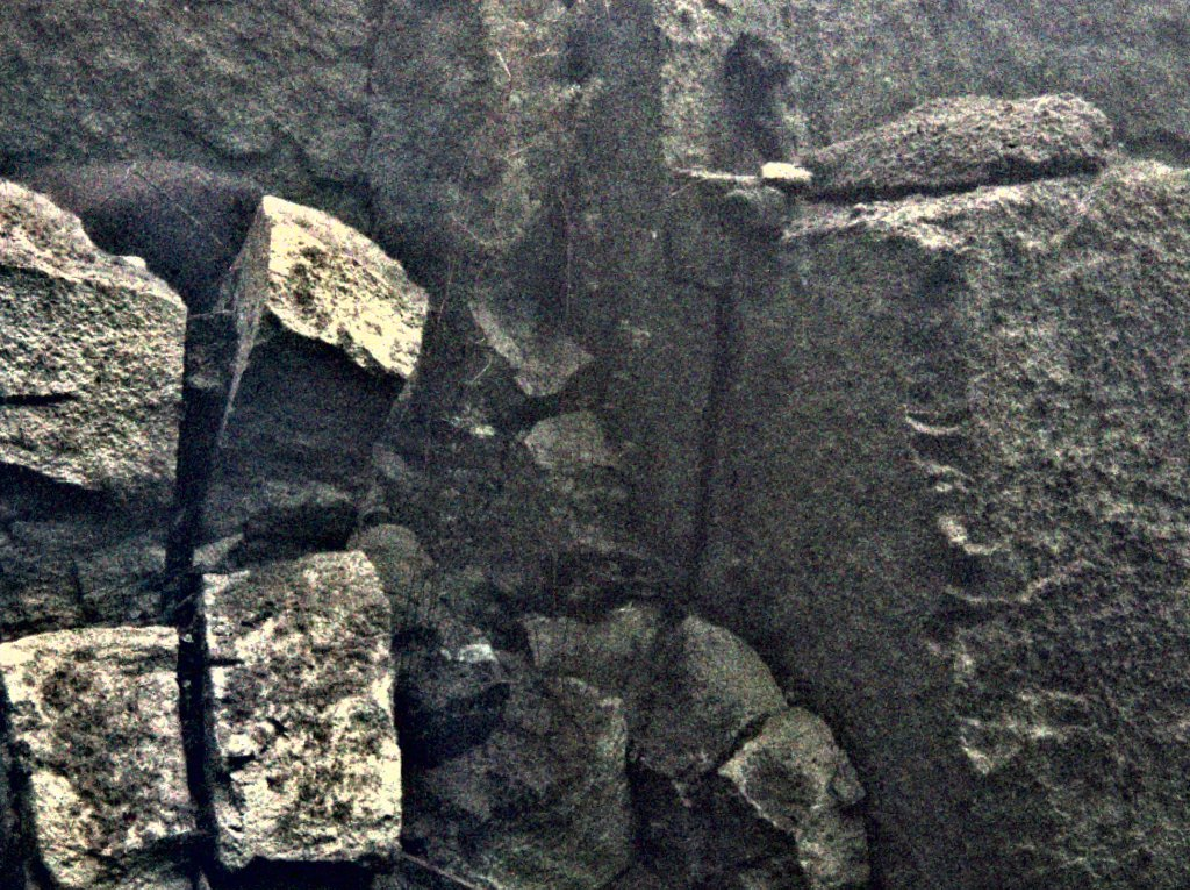}
        \caption{Water filled quarry}
    \end{subfigure}
    \hfil
    \begin{subfigure}{0.45\columnwidth}
        \includegraphics[width=\textwidth]{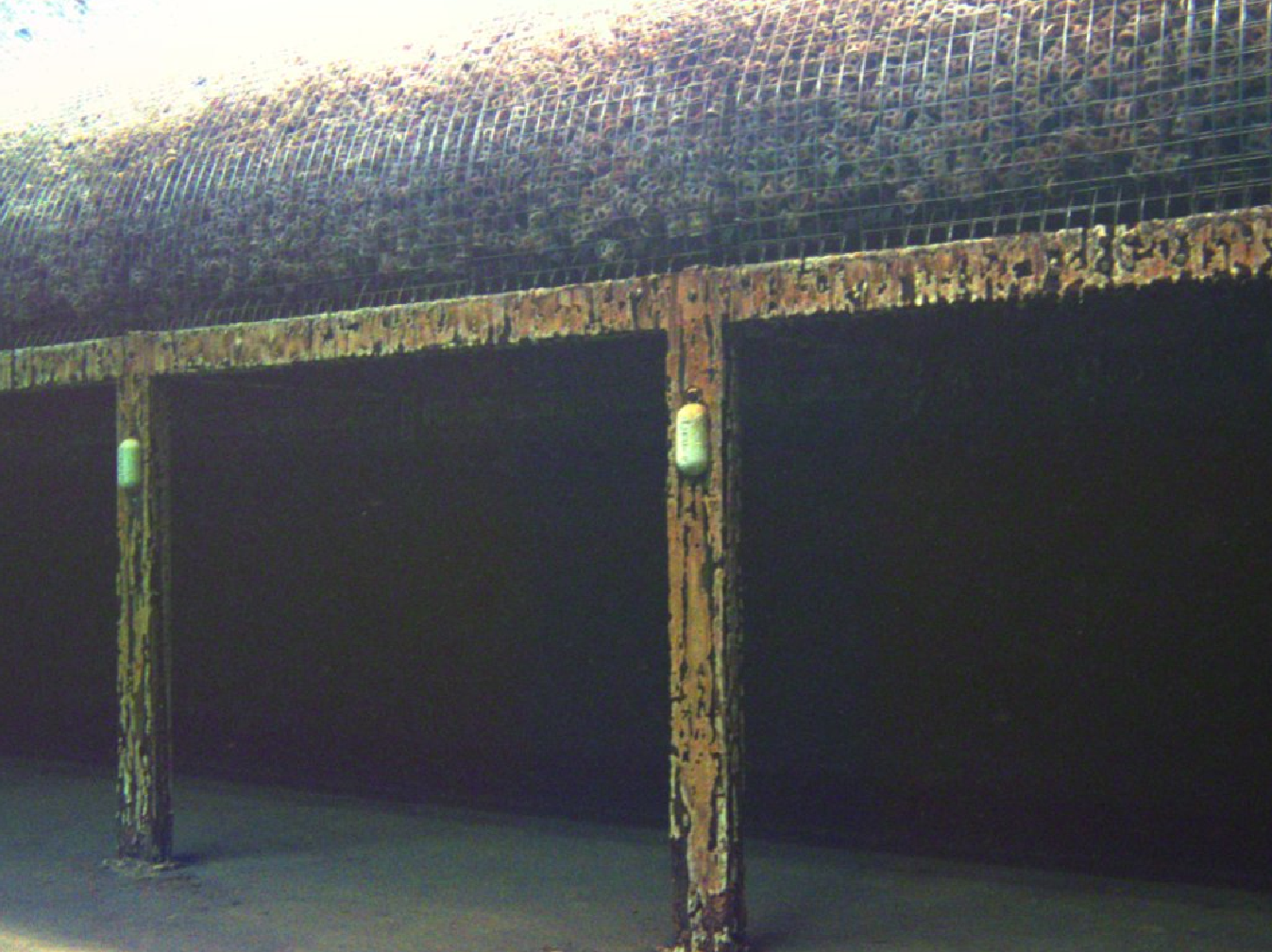}
        \caption{Water tank}
    \end{subfigure}
    \hfil
    \caption{View of the different environments used for the evaluation.}
    \label{fig:eval:env}
\end{figure}

InsSo3D have been tested using a \textit{WaterLinked Sonar3D-15} sonar running at 1.2 MHz with a 90\textdegree~by 40\textdegree~FOV and a maximal operating range of $15m$. It is important to note that our method is independent of the sensors used and would work effectively with other sources of depth images. The system is installed on a \textit{BlueRov 2 Heavy} equipped with a stereo camera \cite{luczynski2019model,9705798} from \textit{Frontier Robotics} and a \textit{Nortek Nucleus 1000} DVL that also includes a built-in AHRS and pressure sensor. The odometry is estimated using \textit{robot\textunderscore localisation} \cite{MooreStouchKeneralizedEkf2014}.

We evaluated InsSo3D in two different environments: i) in a water tank, ii) in a water-filled quarry. Fig. \ref{fig:eval:env} shows the different environments used. Excellent visibility conditions during our trials have enabled us to generate a reference trajectory and 3D geometry of the scene using photogrammetry, which allows us to assess the performance of our method. The photogrammetry reconstruction was performed using the visual-based SfM software \textit{Colmap} \cite{schoenberger2016sfm}, and the scale was estimated based on the stereo camera's baseline. Furthermore, the tank is equipped with a \textit{Qualisys} motion capture system, which was also used to obtain trajectory references during experiments in the test tank. 

\subsection{Evaluation Scenarios}

The first environment evaluated is a water-filled quarry% outside of Edinburgh, UK
. The experiment was conducted at a short operating distance and consisted of $147$ sub-maps during a 47-minute mission over a travelled distance of $230m$. The survey was conducted at short range in order to maintain good visibility conditions for the stereo camera and thus evaluate the quality of the map and trajectory using the SFM results. 

InsSo3D has also been tested in a test tank using a \textit{Qualisys} motion tracking system, allowing for accurate trajectory estimation. Moreover, a map of the test tank has been obtained by photogrammetry, using the images from the stereo camera for map evaluation. The test tank is a difficult evaluation environment for both sonar and compass due to the rectangular shape and the reinforced concrete construction of the tank. This results in multipath effects on the 3D Sonar point cloud, as well as significant drift on the compass and DVL data. This resulted in significant odometry drift during this experiment. In addition, the test tank only has a few geometric features to align with, which makes it a challenging environment for depth based SLAM.

\subsection{Localisation and 3D Reconstruction Accuracy}

\begin{figure}[t]
    \centering
    \begin{subfigure}{0.49\linewidth}
        \includegraphics[width=\textwidth]{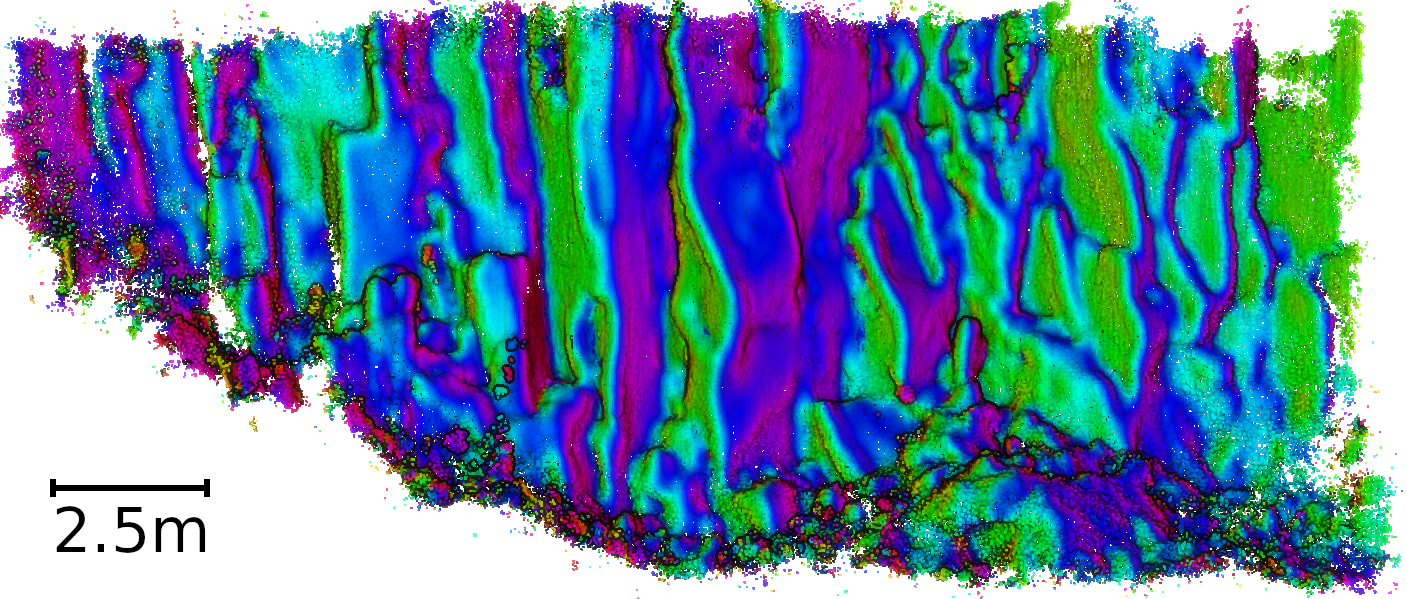}
    \end{subfigure}
    \begin{subfigure}{0.49\linewidth}
        \includegraphics[width=\textwidth]{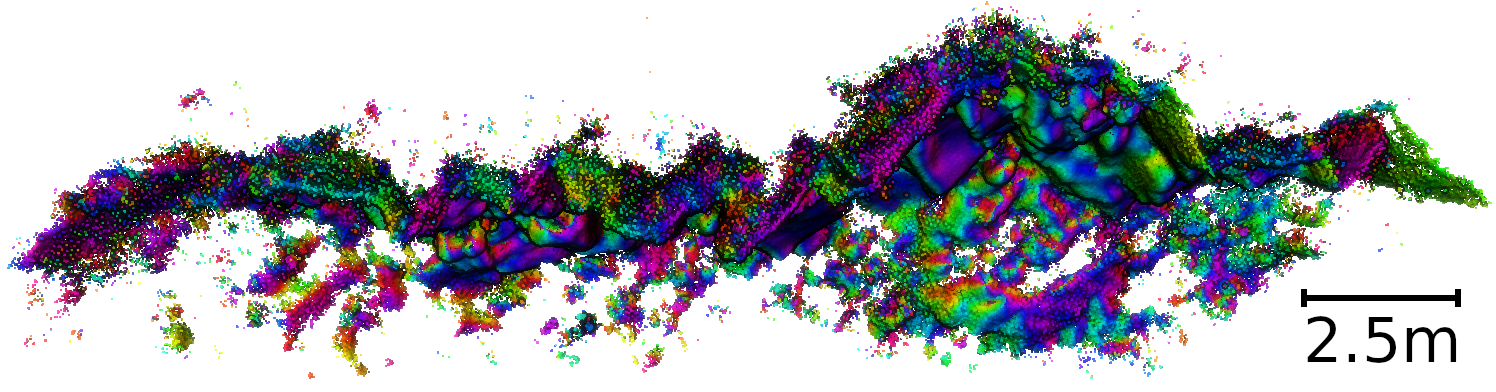}
    \end{subfigure}
    \begin{subfigure}{0.49\linewidth}
        \includegraphics[width=\textwidth]{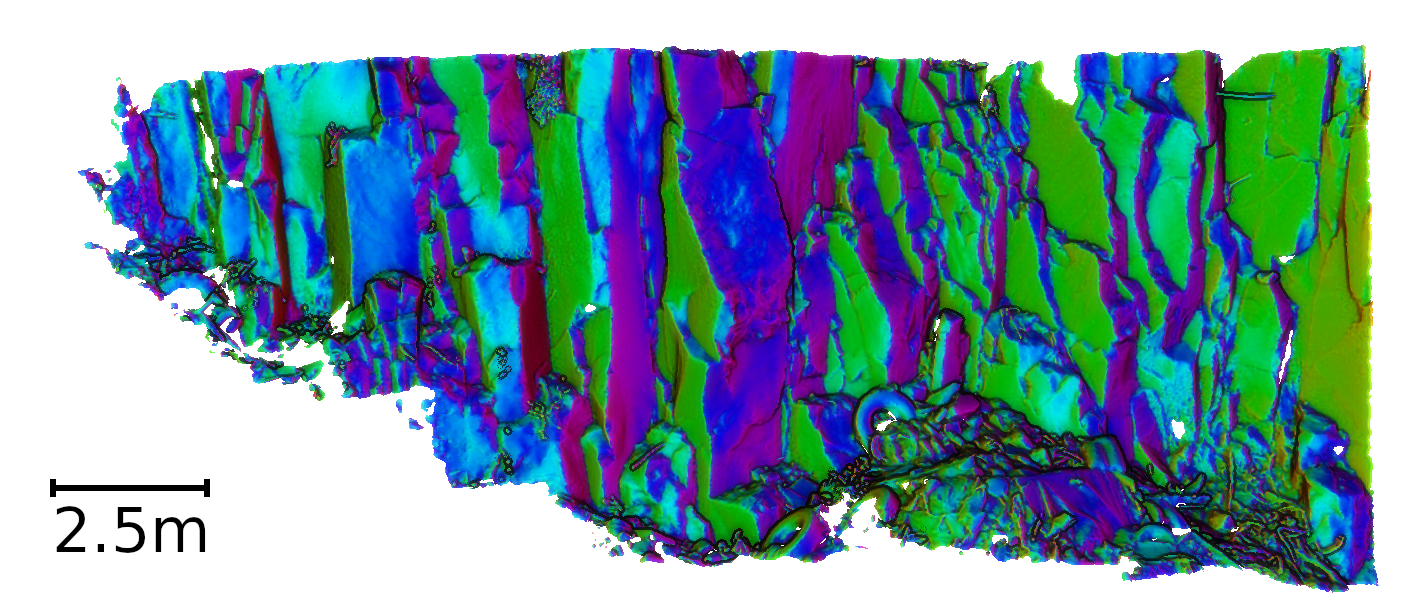}
    \end{subfigure}
    \begin{subfigure}{0.49\linewidth}
        \includegraphics[width=\textwidth]{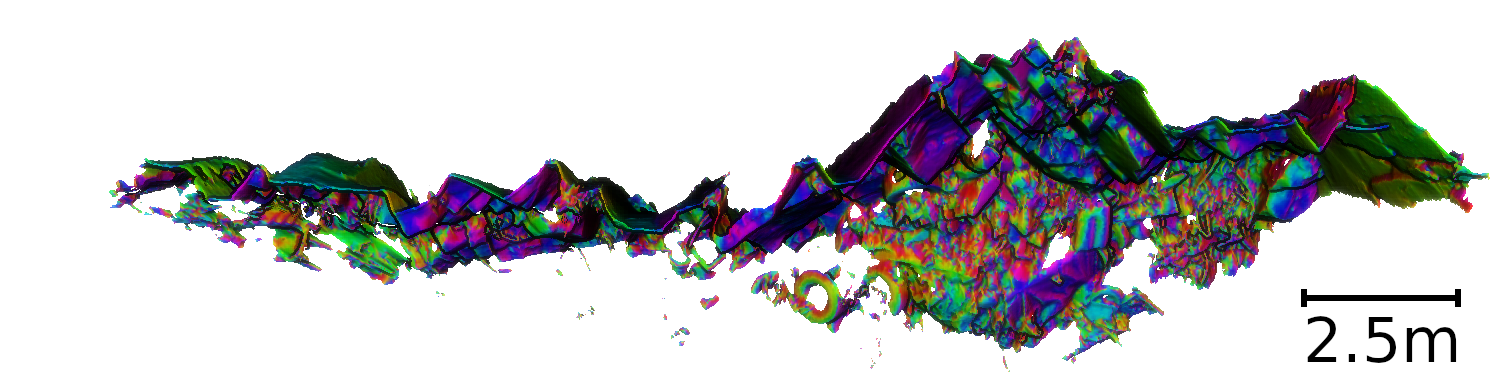}
    \end{subfigure}
    \caption{Map qualitative comparison for the \textit{Quarry} experiments. InsSo3D results are on the first row, and SFM results are on the second row. The first column shows a front view of the maps and the second a top view. The colour represents the normal azimuth angle calculated using the x and y components of the normal vector.}
    \label{fig:eval:map:quarry}
\end{figure}

As the odometry and InsSo3D coordinate systems are different from the reference coordinate system, we first put all trajectories in the same coordinate system with a common origin. Then, the absolute position error $APE(i)$ with respect to the reference throughout the acquisition of 3D Sonar frames is computed. For each trajectory, we also compute the root-mean-square (RMS) and standard deviation (STD) of the APE, denoted $APE_{RMS}$ and $APE_{STD}$. Second, the consistency of the output trajectories of InsSo3D is evaluated, the trajectories are aligned to the references using Umeyama alignment \cite{Umeyama1991}, and the RMS and STD of the resulting absolute position error $APE_{RMS,align}$, and $APE_{STD,align}$ are computed. Finally, we evaluate the output 3D reconstructions by using the mean distance to the reference mesh $e_{map}$ as well as its standard deviation $e_{STD,map}$ after alignment, using CloudCompare \cite{cloudcompare}.

\begin{figure}[t]
    \centering
    \begin{subfigure}{0.49\linewidth}
        \includegraphics[width=\textwidth]{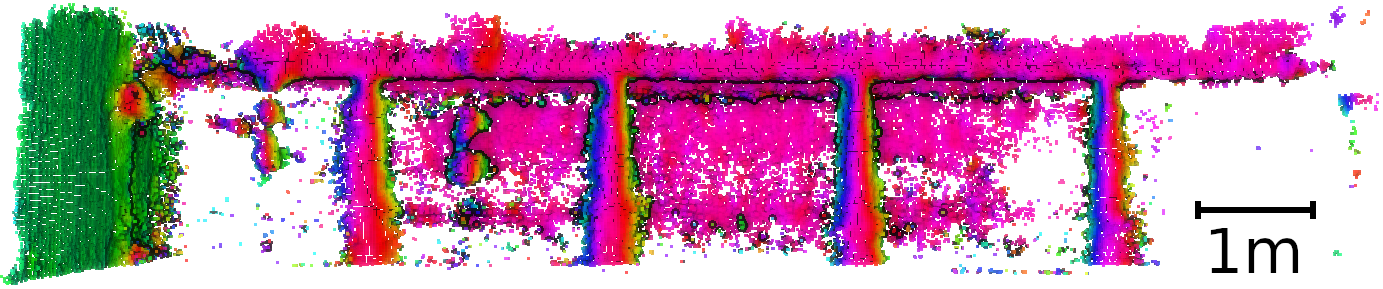}
    \end{subfigure}
    \begin{subfigure}{0.49\linewidth}
        \includegraphics[width=\textwidth]{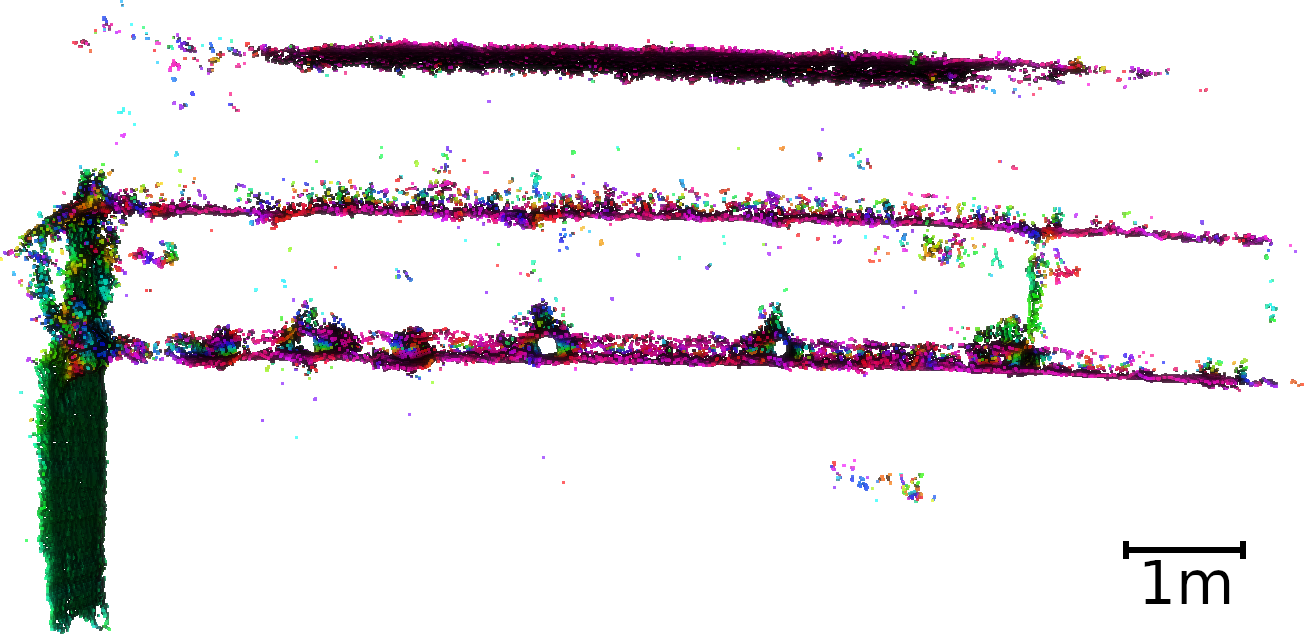}
    \end{subfigure}
    \begin{subfigure}{0.49\linewidth}
        \includegraphics[width=\textwidth]{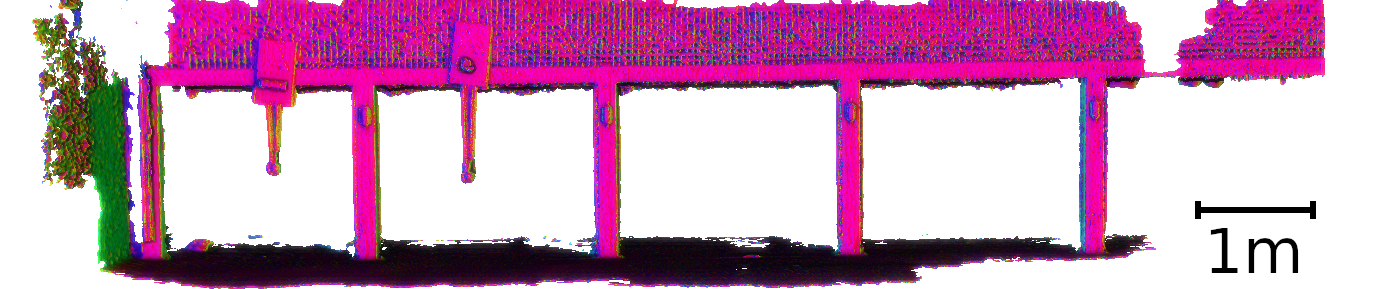}
    \end{subfigure}
    \begin{subfigure}{0.49\linewidth}
        \includegraphics[width=\textwidth]{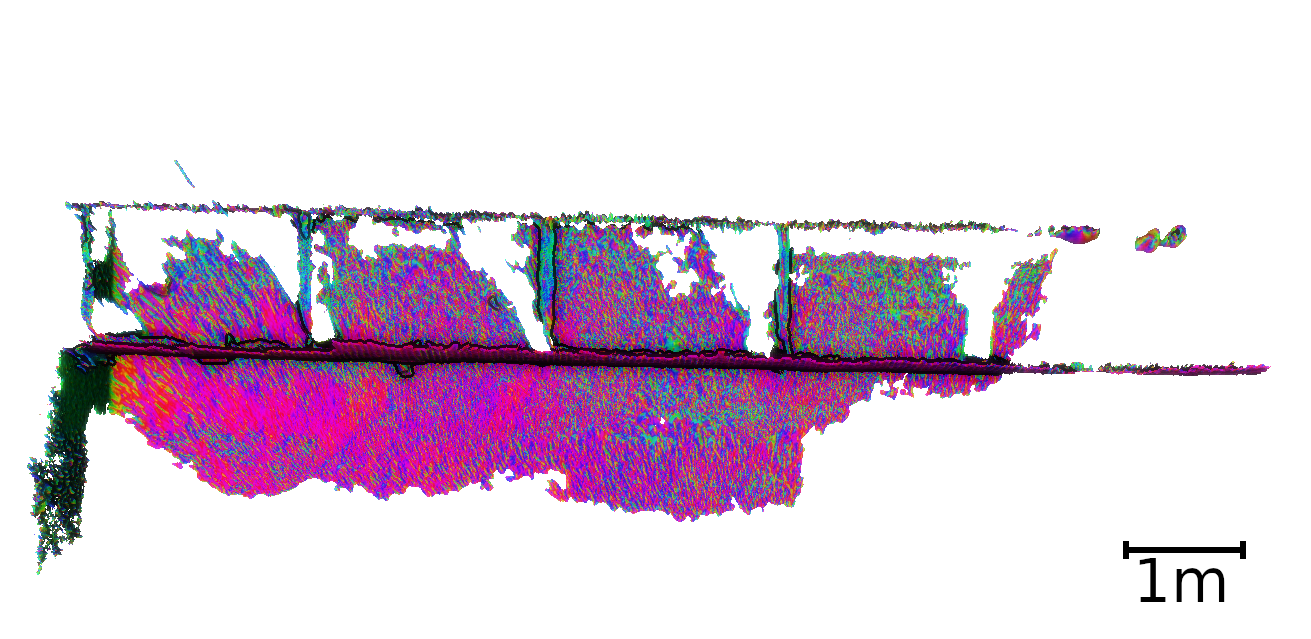}
    \end{subfigure}
    \caption{Map qualitative comparison for the \textit{Tank} experiments. InsSo3D results are on the first row, and SFM results are on the second row. The first column shows a front view of the maps and the second a top view. The colour represents the normal azimuth angle calculated using the x and y components of the normal vector.}
    \label{fig:eval:map:tank}
\end{figure}

We first provide an overview of the maps generated by InsSo3D and compare them with the reference maps of each scenario. Fig. \ref{fig:eval:map:quarry} shows InsSo3D's map reconstruction of the \textit{Quarry} Test environments, and the corresponding SFM results. It can be seen that InsSo3D is able to reconstruct the map with high fidelity, matching closely that of the SFM reference. Figures~\ref{fig:eval:map:tank} present the \textit{Tank} Test environment reconstruction. Despite this environment's complexity, due to sonar reflections and compass issues, InsSo3D still achieves accurate 3D scene reconstruction. Furthermore, it should be noted that thanks to the long range of the 3D Sonar compared to the camera, the back wall of the tank is visible on the InsSo3D map but not on the SFM results.

In all scenarios, InsSo3D provides excellent scene reconstruction, enabling robots to operate in environments where cameras are not suitable, such as low-light or murky water conditions. Additionally, the sonar's larger field of view compared to stereo cameras allows InsSo3D to generate larger maps than SFM, as demonstrated in the \textit{Tank} scenario (Fig.~\ref{fig:eval:map:tank}).

\begin{table}[b]
    \centering
    \begin{tabular}{|c|c||c|c|} 
    \hline
    \multicolumn{2}{|c||}{Sequence} & \textit{Quarry} & \textit{Tank}  \\
    \hline
    \hline
    \multirow{4}{*}{Odom} & $APE_{RMS}$ [m] & 0.942 &  1.784 \\
                          & $APE_{STD}$ [m] & 0.382 &  1.021 \\
                          \cline{2-4}
                          & $APE_{RMS,align}$ [m] & 0.365 &  1.034 \\
                          & $APE_{STD,align}$ [m] & 0.185 & 0.430 \\
    \hline
    \hline
    \multirow{4}{*}{InsSo3D} & $APE_{RMS}$ [m] & \textbf{0.360}  & \textbf{0.089} \\
                          & $APE_{STD}$ [m] & \textbf{0.162}  & \textbf{0.046} \\
                          \cline{2-4}
                          & $APE_{RMS,align}$ [m] & \textbf{0.213} & \textbf{0.078}  \\
                          & $APE_{STD,align}$ [m] & \textbf{0.090} & \textbf{0.045} \\
    \hline
    \hline
    \multicolumn{2}{|c||}{$e_{map}$ [m]} & 0.086 & 0.073 \\
    \multicolumn{2}{|c||}{$e_{STD,map}$ [m]} & 0.0787 & 0.063 \\
    \hline
    \end{tabular}
    \caption{Error and standard deviation of odometry and InsSo3D for trajectory and map (smallest in bold).}
    \label{tab:eval:errors}
\end{table}

\begin{figure*}
    \centering
    \begin{subfigure}{0.49\linewidth}
        \centering
        \begin{tikzpicture}
    \begin{axis}[
        at={(0,0)},
        ylabel={$e_x$ [m]},
        xmin=0, xmax=2730,
        ymin=0, ymax=1.62,
        xticklabels = {},
        legend pos=north west,
        xmajorgrids=true,
        ymajorgrids=true,
        minor tick num=4,
        xminorgrids=true,
        yminorgrids=true,
        minor grid style=dotted,
        width=\linewidth,
        height=2.8cm,
        legend style={at={(0.2,1.1)}, anchor=south},
        legend columns=3,
        title={}
    ]
        \addplot[line width=1pt,solid,color=blue, join=round] table[x=t0, y=ex, col sep=comma] {graph/quarry/odom_error.csv};
        \addplot[line width=1pt,solid,color=purple, join=round] table[x=t0, y=ex, col sep=comma] {graph/quarry/sonar_error.csv};
        \legend{Odom, InsSo3D}
    \end{axis}
    \begin{axis}[
        at={(0,-1.5cm)},
        ylabel={$e_y$ [m]},
        xmin=0, xmax=2730,
        ymin=0, ymax=1.77,
        xticklabels = {},
        xmajorgrids=true,
        ymajorgrids=true,
        minor tick num=4,
        xminorgrids=true,
        yminorgrids=true,
        minor grid style=dotted,
        width=\linewidth,
        height=2.8cm,
    ]
        \addplot[line width=1pt,solid,color=blue, join=round] table[x=t0, y=ey, col sep=comma] {graph/quarry/odom_error.csv};
        \addplot[line width=1pt,solid,color=purple, join=round] table[x=t0, y=ey, col sep=comma] {graph/quarry/sonar_error.csv};
    \end{axis}
    \begin{axis}[
        at={(0,-3cm)},
        ylabel={$e_{yaw}$ [\textdegree]},
        xmin=0, xmax=2730,
        ymin=0, ymax=12.8,
        xticklabels = {},
        xmajorgrids=true,
        ymajorgrids=true,
        minor tick num=4,
        xminorgrids=true,
        yminorgrids=true,
        minor grid style=dotted,
        width=\linewidth,
        height=2.8cm,
    ]
        \addplot[line width=1pt,solid,color=blue, join=round] table[x=t0, y=eyaw, col sep=comma] {graph/quarry/odom_error.csv};
        \addplot[line width=1pt,solid,color=purple, join=round] table[x=t0, y=eyaw, col sep=comma] {graph/quarry/sonar_error.csv};
    \end{axis}
    \begin{axis}[
        at={(0,-4.5cm)},
        ylabel={$APE$ [m]},
        xlabel={$t$ [s]},
        xmin=0, xmax=2730,
        ymin=0, ymax=1.84,
        xmajorgrids=true,
        ymajorgrids=true,
        minor tick num=4,
        xminorgrids=true,
        yminorgrids=true,
        minor grid style=dotted,
        width=\linewidth,
        height=2.8cm,
    ]
        \addplot[line width=1pt,solid,color=blue, join=round] table[x=t0, y=ape, col sep=comma] {graph/quarry/odom_error.csv};
        \addplot[line width=1pt,solid,color=purple, join=round] table[x=t0, y=ape, col sep=comma] {graph/quarry/sonar_error.csv};
    \end{axis}
\end{tikzpicture}
        \vspace{-5mm}
        \caption{\textit{Quarry}}
    \end{subfigure}
    \begin{subfigure}{0.49\linewidth}
        \centering
        \begin{tikzpicture}
    \begin{axis}[
        at={(0,0)},
        ylabel={$e_x$ [m]},
        xmin=0, xmax=143,
        ymin=0, ymax=0.89,
        xticklabels = {},
        legend pos=north west,
        xmajorgrids=true,
        ymajorgrids=true,
        minor tick num=4,
        xminorgrids=true,
        yminorgrids=true,
        minor grid style=dotted,
        width=\linewidth,
        height=2.8cm,
        legend style={at={(0.2,1.1)}, anchor=south},
        legend columns=3,
        title={}
    ]
        \addplot[line width=1pt,solid,color=blue, join=round] table[x=t, y=ex, col sep=comma] {graph/tank/odom_error.csv};
        \addplot[line width=1pt,solid,color=purple, join=round] table[x=t, y=ex, col sep=comma] {graph/tank/sonar_error.csv};
    \end{axis}
    \begin{axis}[
        at={(0,-1.5cm)},
        ylabel={$e_y$ [m]},
        xmin=0, xmax=143,
        ymin=0, ymax=2.86,
        xticklabels = {},
        xmajorgrids=true,
        ymajorgrids=true,
        minor tick num=4,
        xminorgrids=true,
        yminorgrids=true,
        minor grid style=dotted,
        width=\linewidth,
        height=2.8cm,
    ]
        \addplot[line width=1pt,solid,color=blue, join=round] table[x=t, y=ey, col sep=comma] {graph/tank/odom_error.csv};
        \addplot[line width=1pt,solid,color=purple, join=round] table[x=t, y=ey, col sep=comma] {graph/tank/sonar_error.csv};
    \end{axis}
    \begin{axis}[
        at={(0,-3cm)},
        ylabel={$e_{yaw}$ [\textdegree]},
        xmin=0, xmax=143,
        ymin=0, ymax=26,
        xticklabels = {},
        xmajorgrids=true,
        ymajorgrids=true,
        minor tick num=4,
        xminorgrids=true,
        yminorgrids=true,
        minor grid style=dotted,
        width=\linewidth,
        height=2.8cm,
    ]
        \addplot[line width=1pt,solid,color=blue, join=round] table[x=t, y=eyaw, col sep=comma] {graph/tank/odom_error.csv};
        \addplot[line width=1pt,solid,color=purple, join=round] table[x=t, y=eyaw, col sep=comma] {graph/tank/sonar_error.csv};
    \end{axis}
    \begin{axis}[
        at={(0,-4.5cm)},
        ylabel={$APE$ [m]},
        xlabel={$t$ [s]},
        xmin=0, xmax=143,
        ymin=0, ymax=3.1,
        xmajorgrids=true,
        ymajorgrids=true,
        minor tick num=4,
        xminorgrids=true,
        yminorgrids=true,
        minor grid style=dotted,
        width=\linewidth,
        height=2.8cm,
    ]
        \addplot[line width=1pt,solid,color=blue, join=round] table[x=t, y=ape, col sep=comma] {graph/tank/odom_error.csv};
        \addplot[line width=1pt,solid,color=purple, join=round] table[x=t, y=ape, col sep=comma] {graph/tank/sonar_error.csv};
    \end{axis}
\end{tikzpicture}
        \vspace{-5mm}
        \caption{\textit{Tank}}
    \end{subfigure}
    \caption{Position error over $x$, $y$, $yaw$ in and the absolute pose error $APE$ for the different experiments over time $t$.}
    \label{fig:traj-eval}
\end{figure*}
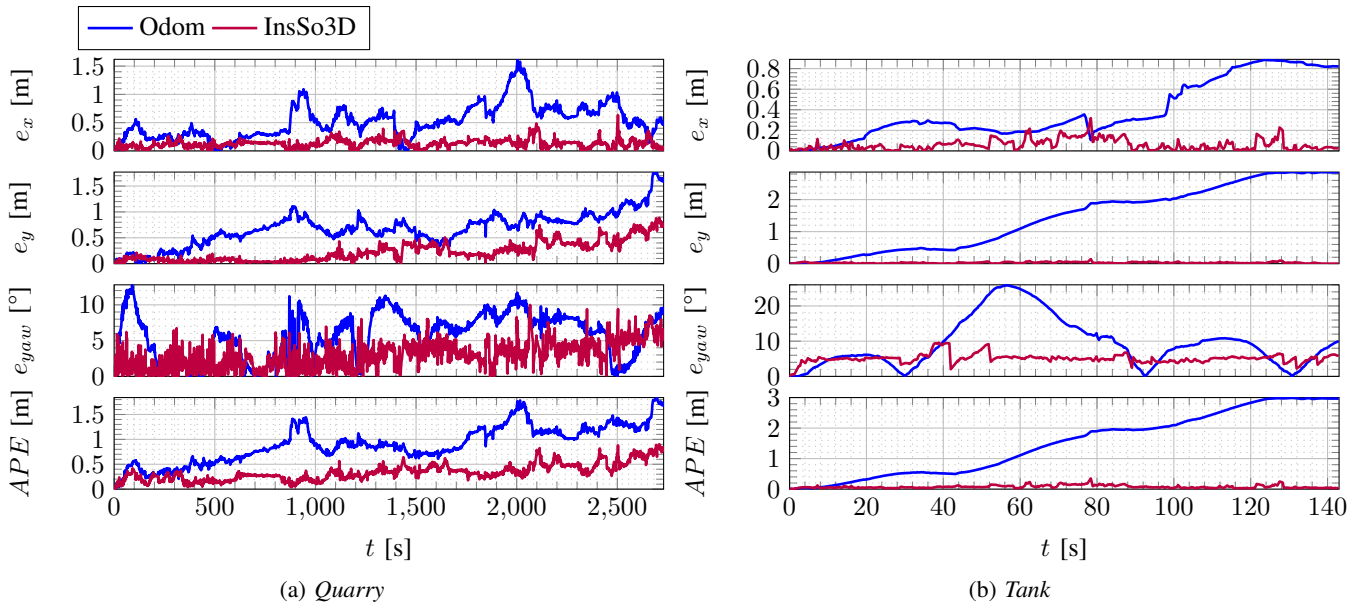

Table \ref{tab:eval:errors} shows InsSo3D's trajectories and map errors, and standard deviation for all scenes. It can be seen that map errors are only a few centimetres for all sequences.
Such small errors can provide very accurate maps, which can be used for inspection or mapping purposes. 

Furthermore, the trajectory errors are given in Fig. \ref{fig:traj-eval}. In all experiments, InsSo3D reduced position drift and standard deviation compared to odometry, even during the long 47-minute missions. In addition, Table \ref{tab:eval:errors} shows that, $APE_{RMS,align}$ is lower than $APE_{RMS}$ for all scenarios, maintaining a very low $APE_{STD,align}$ value, meaning that the trajectories are even more accurate locally, with respect to the scene, than they are with respect to the vehicle starting point. Figure \ref{fig:traj-eval} shows that InsSo3D error remains relatively small, unlike the odometry error. Moreover, during the \textit{Tank} scenario, the odometry drifts significantly by nearly 40 degrees on yaw due to magnetic interference from the reinforced concrete structure of the tank, which disturbs the compass. Even in these situations, InsSo3D is still able to correct its trajectory and maintains the yaw error low. 

Eventually, in all scenarios, InsSo3D $APE_{RMS}$ remains small compared to motion and mapped object scale, and therefore appears relevant for safe navigation in these test scenarios.

\subsection{Runtime Evaluation}

This evaluation considers the processing time of InsSo3D's frontend, backend and global TSDF map generation. The TSDF pipeline is implemented on the GPU using CUDA, while the frontend and backend are CPU-based. The evaluation is performed on the \textit{Quarry} scenario using an \textit{Nvidia RTX 3070} GPU with 8GB of VRAM and an \textit{Intel Core i7-12700H} with 32GB of RAM. Table \ref{tab:runtime} reports the time of the different components. It should be noted that each of these components runs asynchronously; therefore, only the frontend performance impacts the SLAM ability to operate in real time. On this dataset, the frontend takes an average of 70ms per frame, or approximately 14Hz, which is faster than the sonar operating frequency at 6Hz.

\begin{table}[]
    \centering
    \begin{tabular}{|c||c|c|c|}
    \hline
        & Frontend & Backend & Global TSDF  \\
    \hline
         Mean Time [ms] & 69.9 & 593.0 & 4797.4 \\
    \hline
         STD [ms] & 24.3 & 394.3 & 3519.8 \\
    \hline
    \end{tabular}
    \caption{Runtime of InsSo3D main components.The frontend time is reported per frame, the backend is performed per sub-map ($>100$ frames) and the TSDF global map update performed continuously.}
    \label{tab:runtime}
\end{table}

\section{CONCLUSIONS}

This work focuses on underwater online localisation and 3D mapping using a 3D Sonar combined with the vehicle's INS.
We developed a new SLAM framework using frontend and backend graph optimisation to perform robust and large-scale 3D SLAM using a 3D Sonar.
Our comprehensive evaluation, in extensive field experiments, demonstrates that InsSo3D significantly reduces navigation drift while generating high-fidelity 3D maps even during long missions. Comparison with ground truth trajectories and maps shows that InsSo3D is able to achieve trajectory accuracies consistently below $21cm$ and successfully reconstruct quality maps of large-scale environments.
These results establish InsSo3D's effectiveness for real-time, long-range acoustic SLAM in challenging underwater conditions. InsSo3D can be used in real-world deployment for initial map generation in poor visibility conditions, as well as safe motion planning for additional closer inspection. 

While the results obtained from the developed method are promising, there are still limitations that could be addressed in future work. Just like all feature matching algorithms, GICP registration is not robust to environments that contain spatial symmetry or that lack geometric features. Also, in environments prone to sonar multipath effects, the 3D Sonar can produce noisy data and therefore affect the quality of the mapping and localisation. Future work will investigate the fusion of 3D Sonar and camera data to improve robustness and accuracy. Moreover, the camera could also be used to texture the generated map, as well as to improve the localisation.

%\section*{ACKNOWLEDGMENT}
%This work was funded as part of the UNITE EPSRC project EP/XO24806/1 and the AWARE project by NZTC contract SPARK-2660.


%%%%%%%%%%%%%%%%%%%%%%%%%%%%%%%%%%%%%%%%%%%%%%%%%%%%%%%%%%%%%%%%%%%%%%%%%%%%%%%%

%%%%%%%%%%%%%%%%%%%%%%%%%%%%%%%%%%%%%%%%%%%%%%%%%%%%%%%%%%%%%%%%%%%%%%%%%%%%%%%%

%%%%%%%%%%%%%%%%%%%%%%%%%%%%%%%%%%%%%%%%%%%%%%%%%%%%%%%%%%%%%%%%%%%%%%%%%%%%%%%%
%\section*{ACKNOWLEDGMENT}
%This work was funded as part of the UNITE EPSRC project EP/XO24806/1 and the AWARE project by NZTC contract SPARK-2660.

%%%%%%%%%%%%%%%%%%%%%%%%%%%%%%%%%%%%%%%%%%%%%%%%%%%%%%%%%%%%%%%%%%%%%%%%%%%%%%%%

% \clearpage

\bibliographystyle{IEEEtran}
\bibliography{IEEEabrv,biblio}

\end{document}